# ROAD SURFACE TRANSLATION UNDER SNOW-COVERED AND SEMANTIC SEGMENTATION FOR SNOW HAZARD INDEX

*Takato Yasuno, Junichiro Fujii, Hiroaki Sugawara, Masazumi Amakata*

Yachiyo Engineering, Co.,Ltd. RIIPS

**ABSTRACT**

*In 2020, there was a record heavy snowfall owing to climate change. In reality, 2,000 vehicles were stuck on the highway for three days. Because of the freezing of the road surface, 10 vehicles had a billiard accident. Road managers are required to provide indicators to alert drivers regarding snow cover at hazardous locations. This study proposes a deep learning application with live image post-processing to automatically calculate a snow hazard ratio indicator. First, the road surface hidden under snow is translated using a generative adversarial network, pix2pix. Second, snow-covered and road surface classes are detected by semantic segmentation using DeepLabv3+ with MobileNet as a backbone. Based on these trained networks, we automatically compute the road to snow rate hazard index, indicating the amount of snow covered on the road surface. We demonstrate the applied results to 1,155 live snow images of the cold region in Japan. We mention the usefulness and the practical robustness of our study.*

## 1. INTRODUCTION

**1.1. Background and related works**

In Japan, approximately 60 % of land is legally designated as the snow area and cold area, and approximately 20% of people have inhabited regions such as the Hokkaido, North Tohoku, Hokuriku, and Sanin. In the past two decades, approximately 4,000 traffic accidents have occurred per year during winter [1]. The primary reason includes slippery road (91%), reduced ability to see (6%), and rutted road (2%). In winter, following road conditions triggered accidents: iced (50%), dried (29%), wet (11%), and snow (10%). There are several primary hazards of winter driving [2]. They are: 1) poor traction, 2) reduced ability to stop vehicles, 3) the effect of temperature on starting and stopping vehicles, 4) slippery road conditions owing to ice and snow, 5) reduced ability to see and be seen, and 6) jackknifing. Recently, on January 10, 2021, owing to heavy snowfall on the Hokuriku Expressway, approximately 1,000 cars were stuck on the up and down lines between the Fukui and Kanazu interchange. This congestion was caused by a slippery accident. Therefore, road surface monitoring is essential for road safety.

In the past decade, the focus has always been on an intelligent system for monitoring winter road surface conditions via computer vision and machine learning [3–14]. Takahashi et al. studied a prediction method for winter road surfaces corresponding to the change in snow removal levels [4]. Further, the road surface temperature and condition were estimated based on field observations. However, this method required several observed weather inputs such as temperature, wind velocity, humidity, sky radiation, and sunshine and incurred a higher cost for utilizing multi-mode data. Grabowski et al. proposed an intelligent road sign system including a detector using convolutional neural networks (CNNs) and image processing video cameras for the classification of dry, wet, and snowy roads [13]. Liang et al. investigated the application of semantic segmentation network customized with dilated convolutions for road surface status recognition [14]. The focus has been on performance improvement for road surface recognition; however, an important post-process of providing hazardous information on the winter road surface condition for road managers and users still remains.

**1.2. Monitoring road surface for snow hazard alert**

As illustrated in Fig. 1, we propose an application to provide snow hazard index information with snow hazard alerts using live camera images. If drivers can acquire information on snow levels through live cameras before their travel, they can decide to use snow adaptive tires to avoid tripping, in case of wet and slippery road conditions. The increasing usage of CCTV cameras has opened the possibility of using them to automatically detect hazardous road surfaces and inform drivers through alerts. In winter, we aim to provide a new snow hazard index on the amount of snow covered on the road surface, whose value is from zero to 100. During a heavy snowfall, the road surface region is covered with snow and is not clearly visible on the live camera. Through live image, the background snow excluded with road region would not influence road user for snow hazard. Thus, we need to know the ratio of snow covered on the road.

This study proposes a deep learning application with live image post-processing to automatically compute a snow hazard indicator. First, the road surface region hidden under snow is translated using a conditional generative adversarial network (GAN). Second, snow-covered and road surface classes are detected using semantic segmentation. Further, the road to snow rate hazard index is automatically computed, indicating the amount of snow covered on the road surface. Finally, the pipeline are applied on the snow and cold regions in Japan.

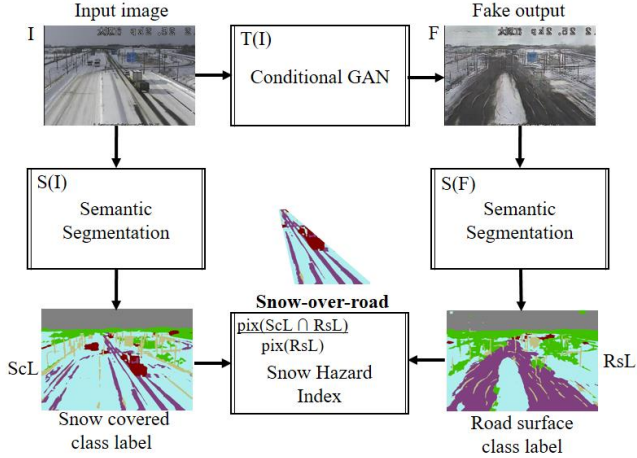

**Fig. 1** Overview of our pipeline showing the raw input is translated by the conditional GAN network T. The resulting road surface fake output F is further fed into the semantic segmentation network S to predict a *road surface class label* output RsL. Simultaneously, the raw input is predicted into a *snow covered class label* output ScL. Finally, the intersection of both the outputs is the dangerous region of interest (ROI) *snow-over-road*. The ratio index based on two counting pixels of ROIs is computed by dividing the snow-over-road ROI by the road surface class ROI.

## 2. DEEP LEARNING APPLICATION

Our proposed application contains the conditional GAN, semantic segmentation, and image processing. We formulate each task as follows.

### 2.1. Conditional GAN for generating road surface region

The image-to-image translation is possible for training a paired image dataset with fixed camera angle. This study proposes a method to generate the road surface from snow-covered region using L1-Conditional GAN. In the original pix2pix study, the input edge images were translated to shoe images [15]. Further, using the Yosemite photo dataset, they translated the summer photo to winter photo. However, in the case of snow-covered road images, it is difficult to detect the road surface region. This study applies a generative deep learning method, in which the road surface region hidden under snow is translated using the conditional GAN, pix2pix.

### 2.2. Semantic segmentation for snow and road classes

There are useful architectures such as semantic segmentation algorithms using deep learning for image analysis. The first end-to-end semantic segmentation neural network is the fully convolutional network (FCN) [16]. Several modifications have been made to FCN to improve performance. SegNet refines the encoder-decoder network layers with skip connections and conv-batchnorm-relu layers [17]. DeepLab replaced the deconvolutional layers with densely convolution and atrous spatial pyramid pooling instead of standard convolution layers [18].

This study applies a semantic segmentation method to detect snow-covered and road surface classes using a DeepLabv3+ algorithm with MobileNet as a backbone. This network has low memory and is practically accurate and fast for live camera monitoring of winter roads.

### 2.3. Image processing for snow hazard index

The input raw image is abbreviated as I. Based on the aforementioned deep learning, two trained networks were used: a conditional GAN network T and a semantic segmentation network S. As depicted in Fig. 2, we propose a pipeline to automatically compute a snow hazard index of the amount of snow covered on the road surface. First, a raw input image is translated into a road surface fake output F = T(I) using the trained pix2pix network T. In addition, the road surface output F is indirectly detected into the *road surface class label* output RsL using the trained semantic segmentation network S. The first image process is represented as two mapping function.

$$T(I) \rightarrow F, \qquad S(F) = S(T(I)) \rightarrow RsL. \quad (1)$$

Thus, we are able to focus on the monitoring scope ROI, which is the road surface region, though it has been hidden on the raw snow-covered input image. Notably, only the raw input image enable to approximate the road surface scope.

Simultaneously, a raw input image is directly detected into the *snow covered class label* output ScL = S(I) using the trained semantic segmentation network S. The second image process is expressed as a next mapping.

$$S(I) \rightarrow ScL. \quad (2)$$

Furthermore, the intersection of both the outputs is highlighted on the dangerous target ROI *snow-over-road*. The ratio index based on two counting pixels of ROIs is computed by dividing the snow-over-road ROI by the road surface class ROI. Thus, a *snow hazard ratio* index, SHR, is formulated as next equations by counting pixels.

$$SHR = pix(RsL \cap ScL)/pix(RsL). \quad (3)$$

Here, pix(L) represents the pixel count of the class label region L. This snow hazard ratio index is represented as the function of the only input image I, conditional to the trained networks such as the conditional GAN translator network T and semantic segmentation network S.

$$SHR(I|T,S) = pix(S(T(I)) \cap S(I))/pix(S(T(I))). \quad (4)$$

Thus, the snow hazard ratio index is automatically computed using only the input raw image, instead of using the ground-truth image of the road surface without snow.

## 3. APPLIED RESULTS

Herein, we demonstrate that our pipeline could automatically compute a snow hazard ratio index at the cold region. Fig. 2 depicts the 16 test images randomly selected. In contrast, Fig. 3 illustrates the ground-truth images of the road surface without snow. Fig. 4 depicts a road surface fake output translated from a raw input using a trained pix2pix.

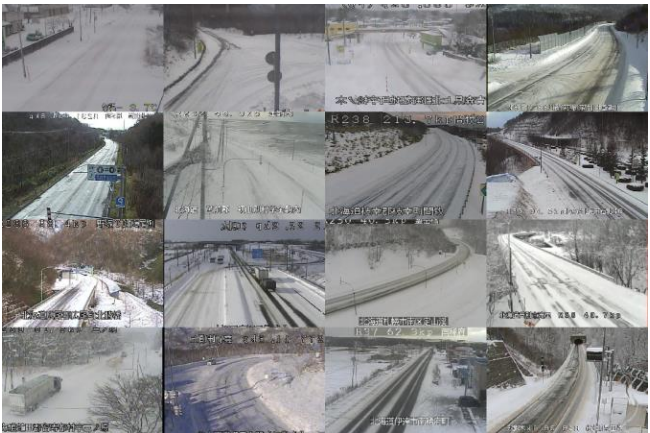

**Fig. 2** Input raw images examples (4 × 4 montage)
From [19] https://www.mlit.go.jp/road/road_e/index_e.html

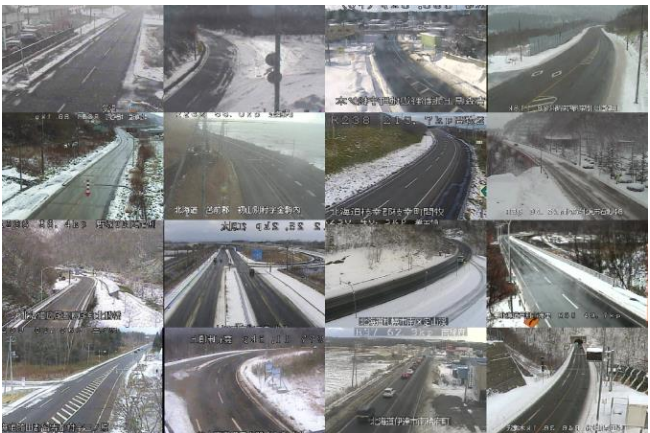

**Fig. 3** Ground-truth road surface examples (4 × 4 montage)
From [19] https://www.mlit.go.jp/road/road_e/index_e.html

### 3.1. Generating road surface from snow-covered regions

We prepared 1,155 one-to-one datasets that included two groups: the snow covered road and the road surface without snow. We collected live road images from December 2 2020 to January 21 2021 at the cold region in Japan [19]. We unified 480 × 720-size images for training. The ratio of training data versus test data was 90:10. We set the input size of pix2pix network as 512 × 768. We iterated 200 epoch using the Adam optimizer; therefore, it took 65 hours.

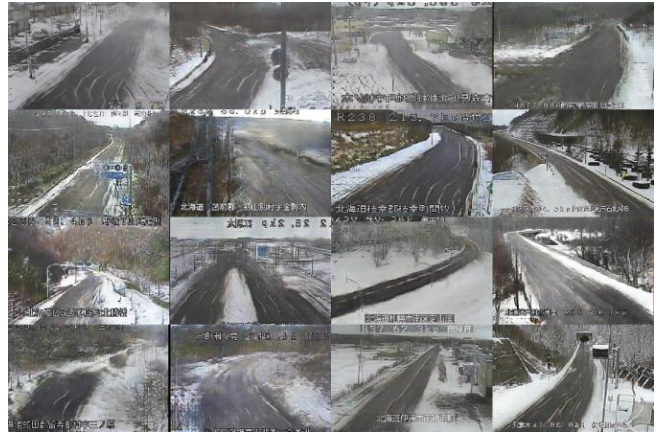

**Fig. 4** Fake output road surface generated from input raw images using pix2pix (4 × 4 montage)

### 3.2. Prediction of snow and road surface segments

We prepared 500 annotated datasets that contained the raw images and six class defined labels such as road, pole-sign, green, snow, sky, and background. We resized images of 598 × 1196 size and extracted 2 × 4 crops, each with size 299 × 299; therefore, the total number of datasets was 4,000. The ratio of training data versus test data was 93:7. We set the input size of a semantic segmentation as 299 × 299. We trained 30 epoch iterations with a mini-batch of 16 and learning rate of 0.001 using the Adam optimizer; therefore, it took 3.5 hours. As presented in Fig. 5, each class mIoU had almost high accuracy scores, with the snow and road classes having the same score of more than 0.8. The entire accuracy mIoU was 0.7718, mean accuracy was 0.8572, and mean F1 score was 0.6886.

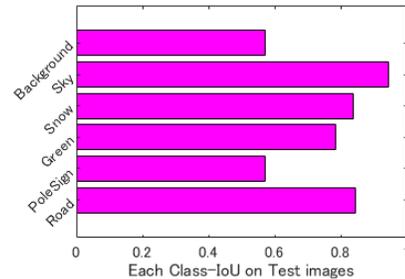

**Fig. 5** Each class mean intersection of union (mIoU) results

As displayed in Fig. 6, the road surface fake output was segmented into the road class label output using the trained network DeepLabv3+ with MobileNet as a backbone. As depicted in Fig. 7, the input image was segmented into the snow class label output using the trained deep network.

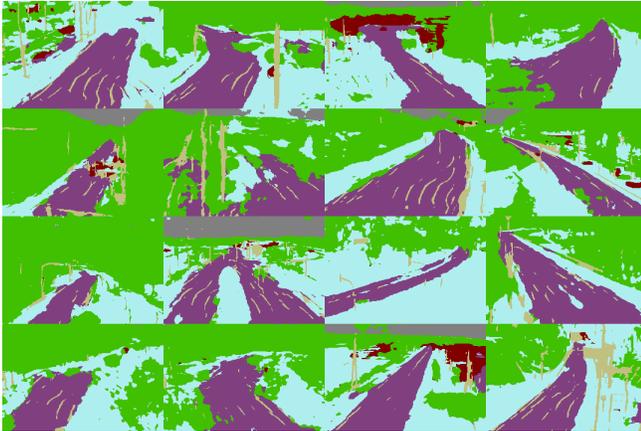

**Fig. 6** Road surface label results predicted from the road surface fake outputs using DeepLabv3+ (4 × 4 montage)

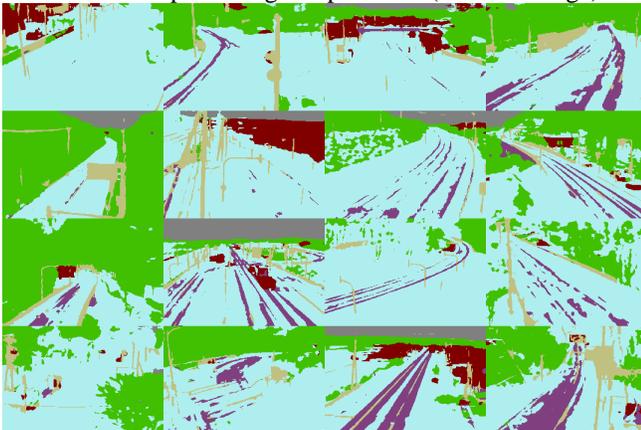

**Fig. 7** Snow region label results predicted from the input raw images using DeepLabv3+ (4 × 4 montage)

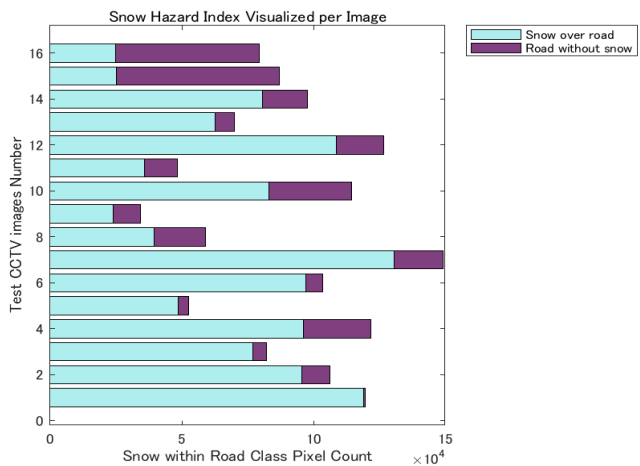

**Fig. 8** Bar plots of the pixel count for the dangerous target ROI snow-over-road and road surface without snow

### 3.3. Computing snow hazard index for test images

Fig. 8 presents the pixel count for 16 image results for the *snow-over-road* (light cyan) and the *road without snow* (purple) ROIs. Each image has different size of road surface ROI. Each snow region does not include the background snow, which never influences the safety of the road user. Fig. 9 presents the applied results of *the snow hazard ratio index*. Thus, it could automatically provide road managers and users with the alert information of snow over the road.

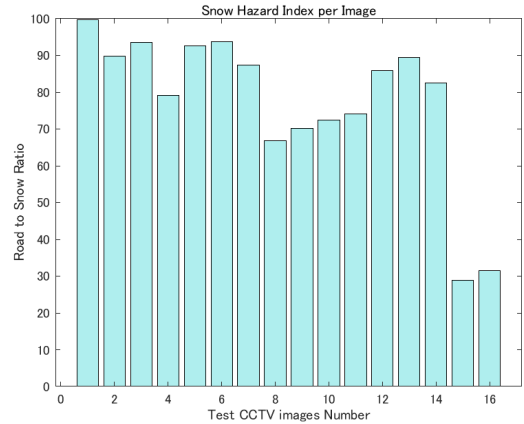

**Fig. 9** Bar plots of the *snow hazard ratio* index; indicates the snow ratio within the road surface ROI of each image.

### 4. CONCLUDING REMARKS

This study proposed a deep learning application with live image to automatically compute a snow hazard ratio indicator. First, the road surface hidden under snow was translated using a conditional GAN, pix2pix. Second, hazardous target ROI snow-over-road and road surface region were segmented using DeepLabv3+ with MobileNet as a backbone. Further, the snow hazard ratio index was computed for the amount of snow covered on the road surface. We demonstrated our pipeline to the cold region in Japan. This application had the advantage of using only one live image as the input, without any mining of before and after paired images dataset. Furthermore, the indicator could be delivered to the road managers and users using our pipeline automatically computed *snow hazard ratio index*, whose value was zero to 100 for multi-points comparison.

In future, the data mining and annotation are required to be conducted to achieve a higher accuracy of pix2pix and semantic segmentation. Future work involves transfer learning for urban traffic in regions with rare snowfall, for example, metropolitan area. Further, there are specific problems for winter road safety. For example, black ice on a seemingly wet road surface and bridge deck icing on the river. Focusing on the critical snow hazard, we continue to collect hazardous feature images and build a classifier for preprocess to detect strong hazard class. Furthermore, we extend our pipeline to edge computing for automatic real-time provision for drivers about hazardous snow points.

**Acknowledgments** I would like to thank Takuji Fukumoto and Shinichi Kuramoto (MathWorks Japan) for providing us MATLAB resources for deep learning frameworks.